\documentclass[letterpaper, 10 pt, conference]{ieeeconf}  

\IEEEoverridecommandlockouts                              

\usepackage{graphics} 

\usepackage{amsmath}

\usepackage{pgfplots}
\pgfplotsset{compat=1.5}

\usepackage{multirow}

\usepackage{url}

\usepackage{hyperref}

\usepackage[noend]{algpseudocode}
\usepackage[ruled,noend]{algorithm2e}

\usepackage{multirow}
\usepackage{array}
\newcolumntype{P}[1]{>{\centering\arraybackslash}p{#1}}
\newcolumntype{M}[1]{>{\centering\arraybackslash}m{#1}}
\newcolumntype{L}[1]{>{\raggedleft\arraybackslash}m{#1}}
\newcolumntype{R}[1]{>{\raggedright\arraybackslash}m{#1}}

\usepackage{amsmath}
\usepackage{color}

\title{\LARGE \bf
Planning High-Quality Grasps using \\Mean Curvature Object Skeletons
}

\author{
Nikolaus Vahrenkamp, Eduard Koch, Mirko W\"achter and Tamim Asfour%
\thanks{The authors are with the High Performance Humanoid Technologies (H\textsuperscript{2}T) lab, Institute for Anthropomatics and Robotics (IAR), Karlsruhe Institute of Technology (KIT), Germany}
\thanks{The research leading to these results has received funding from the European Union’s Horizon 2020 Research and Innovation programme under grant agreement No 643950 (SecondHands).}
}

\begin{document}

\maketitle
\thispagestyle{empty}
\pagestyle{empty}
		
\begin{abstract}
In this work, we present a grasp planner which integrates two sources of information to generate robust grasps for a robotic hand. First, the topological information of the object model is incorporated by building the mean curvature skeleton and segmenting the object accordingly in order to identify object regions which are suitable for applying a grasp.
Second, the local surface structure is investigated to construct feasible and robust grasping poses by aligning the hand according to the local object shape. 
We show how this information can be used to derive different grasping strategies, which also allows to distinguish between precision and power grasps. We applied the approach to a wide variety of object models of the KIT and the YCB real-world object model databases and evaluated the approach with several robotic hands. The results show that the skeleton-based grasp planner is capable to autonomously generate high-quality grasps in an efficient manner. 
In addition, we evaluate how robust the planned grasps are against hand positioning errors as they occur in real-world applications due to perception and actuation inaccuracies. 
The evaluation shows that the majority of the generated grasps are of high quality since they can be successfully applied even when the hand is not exactly positioned.

    \end{abstract}
    

    %
    \IEEEpeerreviewmaketitle

    \section{Introduction}

The ability to robustly plan feasible grasps for arbitrary objects is an important capability of autonomous robots in the context of mobile manipulation. It allows to generate grasping configurations in an autonomous and unsupervised manner based on the object's 3D mesh information.
Part-based grasp planning is an approach to deal with the complexity of arbitrary shaped objects by segmenting them into simpler parts which are more suitable for shape analysis for grasp generation. This procedure complies also with the behavior of humans which prefer to structure objects into smaller segments (see e.g. the recognition-by-components paradigm of Biederman~\cite{biederman1987}).
Representing an object by its parts fits very well to robotic grasp planning, since a robotic gripper or hand can usually only grasp an object by accessing a local surface area for establishing contacts.
\begin{figure}[t]%
\centering
\includegraphics[width=0.9\columnwidth]{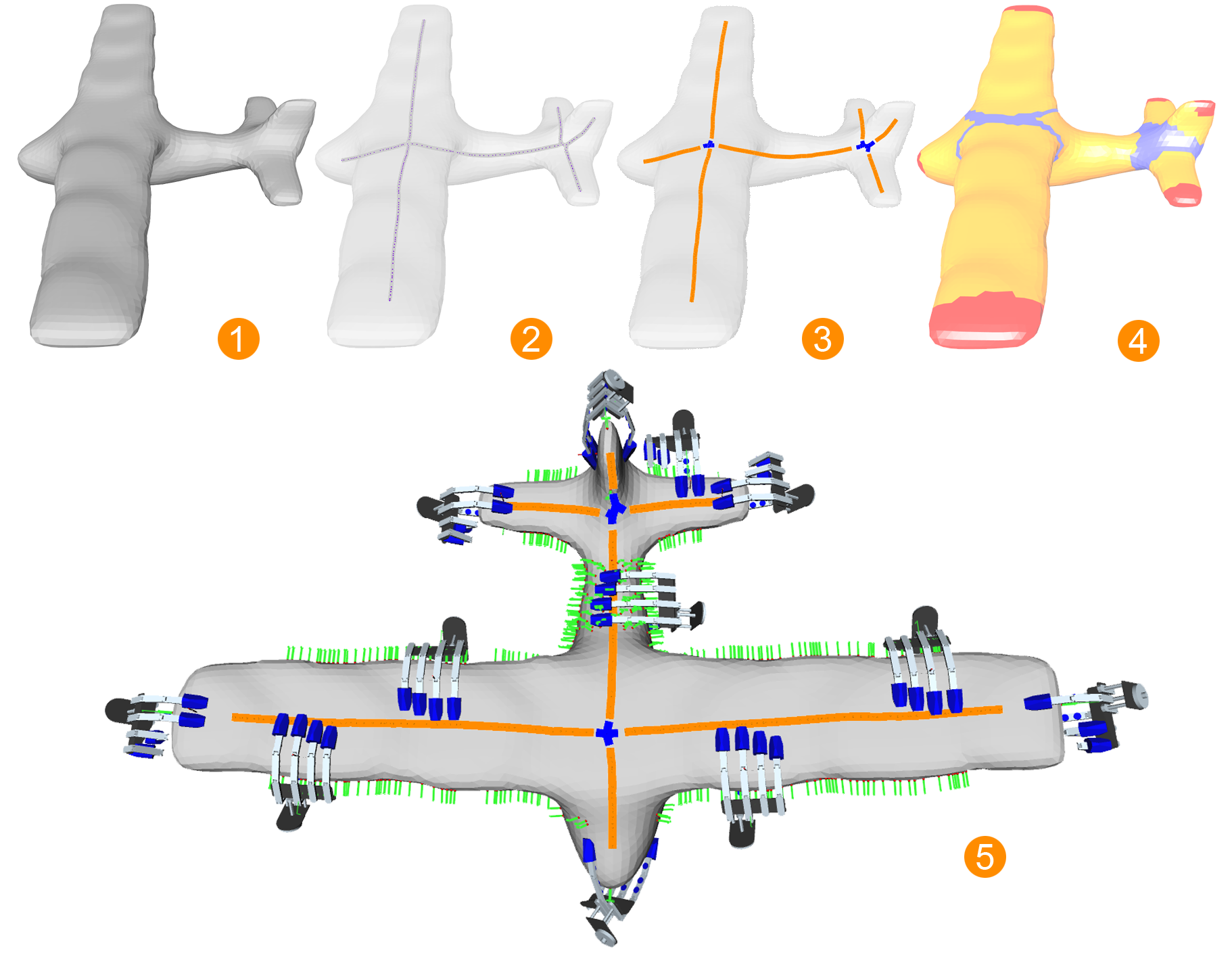}%
\caption{The object mesh model (1) is used to build the mean curvature object skeleton (2). The skeleton is segmented (3) and the corresponding surface segments (4) are depicted in red (segment end points), yellow (connecting segments), and blue (segment branches).
The results of the skeleton-based grasp planner are visualized (5) by indicating the approach directions of the planned grasps via green lines. In addition, a selection of grasps is fully visualized.}
\label{fig:teaser}%
\end{figure}
The part-based grasp planner presented in this work is based on the assumption that an object can be segmented according to its skeleton structure. The object skeleton simplifies the topological representation and provides information about the connectivity of the segments.
Object segmentation is realized by analyzing the skeleton according to branches and crossings in the topology. 
The object segments are analyzed separately to determine if several grasping strategies (e.g. power or precision grasps) can be applied according to the local surface information (see \autoref{fig:teaser}).

Since robustness is essential for robotic grasp planning, it is desirable to plan grasps which can be executed robustly in realistic setups. Hence, we aim at planning grasps that are robust to disturbances and inaccuracies as they appear during execution due to noise in perception and actuation. In this work, we show that the analysis of the object skeleton in combination with the usage of local surface properties leads to grasping hypotheses which are robust to inaccuracies in robot hand positioning, which indicates a reliable execution on real robot platforms. An implementation of the presented grasp planning approach based on the Simox library \cite{Vahrenkamp12b} is provided as C++ open source project\footnote{https://gitlab.com/Simox/simox-cgal}.



    \section{Related Work}
\label{sec-realted-work}

Napier divided hand movements in humans into \textit{prehensive movements} and \textit{non-prehensive movements} to distinguish between suitable and non-suitable grasping movements \cite{napier1956prehensile}. 
He showed that prehensive grasping movements of the hand consist of two basic patterns which he termed precision grasp and power grasp. 
Based on this work Cutkosky developed a grasp taxonomy which distinguishes between 16 different grasping types \cite{cutkosky1989}. 
The transfer of grasping strategies in humans to robotic applications is usually done by reducing the complexity, e.g. by considering a low number of grasp types (i.e. power and precision grasp) or by using the Eigengrasp approach which is capable of approximating human grasping movements with low degree of freedom spaces \cite{CiocarlieA09}.

Approaches for grasp synthesis in the literature are usually divided into analytical and empirical or data-driven algorithms \cite{bohg2014, sahbani2012overview}.
Analytical approaches are based on geometrical properties and/or kinematic or dynamic formulations, whereas data driven approaches rely on simulation. The complexity of the problem is often reduced by utilizing randomized methods and by considering simplified contact models.
Data-driven approaches also make use of simulation environments such as \textit{Graspit!}~\cite{Miller2004}, OpenRave~\cite{Diankov2010}, and Simox~\cite{Vahrenkamp12b} to generate and evaluate grasping poses. Many works generate grasping hypotheses and evaluate them by applying the Grasp Wrench Space approach which allows to determine if a grasp is \textit{force-closure} (i.e. valid) and to compute $\epsilon$ or \textit{Volume} quality values indicating how much the grasping configuration can resist to external forces \cite{ferrari1992planning, Roa2015}.

The execution of such grasps with real robot hands is challenging since small disturbances, as they appear during real-world execution, could quickly lead to unstable grasping configurations as shown in \cite{weisz2012pose}. It has been shown that such potential inaccuracies in execution can be considered during grasp planning by analyzing small variations of the generated grasping pose.

In part-based grasp planning there exists several approaches which consider parts of the object for grasp synthesis.
In \cite{miller2003} the object is approximated by a set of primitive shapes (box, cylinder, sphere, cone) to which manually defined hand configurations can be assigned. 
A similar approach has been presented in \cite{goldfeder2007}. The object is decomposed in superquadrics and several heuristics are used to generate the grasping information. 
In \cite{huebner2008} the object is decomposed based on Minimal Volume Bounding Boxes. grasping information is synthesised on the resulting bounding boxes. An extension of this work is presented in \cite{geidenstam2009} where neural networks are used to combine 3D and 2D grasping strategies. 
The grasp planner presented in \cite{aleotti2010grasp} is operating on a topological object segmentation which is computed based on the Reeb Graph formalism. The resulting object segments are used for randomized grasp generation. 
Independent Contact Regions can be used to determine surface areas that are suitable for grasping \cite{Roa07}.
An voxelized object representation is used by \cite{Song2016} to plan feasible grasps via hand-object geometric fitting strategies. 

Our previous work described in \cite{pryb2010, przybylski2012} is related to the work we present here since it uses the medial axis transform as a simplified object representation on which grasp planning is performed. The medial axis representation is computed on point clouds or mesh data and provides information about the object structure and object symmetries. Before grasp planning can be performed, the medial axis transform is processed via cluster algorithms and transferred to grid-based data structures on which grasp synthesis is realized based on several heuristics for typical grid structures. In contrast to \cite{przybylski2012}, we use mean curvature skeletons \cite{tagliasacchi2012mean} to represent the object structure.  Compared to the medial axis approach, this representation results in a reduced complexity while preserving full object surface information (see Section~\ref{sec-object-skeleton}). We exploit the skeleton structure for object segmentation and in Section~\ref{sec-planning} we show how the object segments can be analyzed for part-based grasp planning. The approach is evaluated in Section~\ref{sec-eval} on several object data sets by investigating the force closure rate and the robustness of the generated grasps according to \cite{weisz2012pose}. In addition, we compare the results to a randomized grasp planning algorithm which aligns the approach direction to the object surface \cite{Vahrenkamp12b}, similar to the approach used in \cite{Diankov2010} and \cite{Kappler2015}.

    \section{Object Skeletons}
\label{sec-object-skeleton}

3D mesh objects are processed in order to generate mean curvature skeletons which provide a medically centered skeleton representing the object's topology \cite{tagliasacchi2012mean}. As we show in this section, the skeleton data structure is used to segment the object according its topology. 

\subsection{Mean Curvature Skeletons}
\label{sec-skeleton}

\begin{figure}[tbh]%
\centering
\includegraphics[height=0.13\textheight]{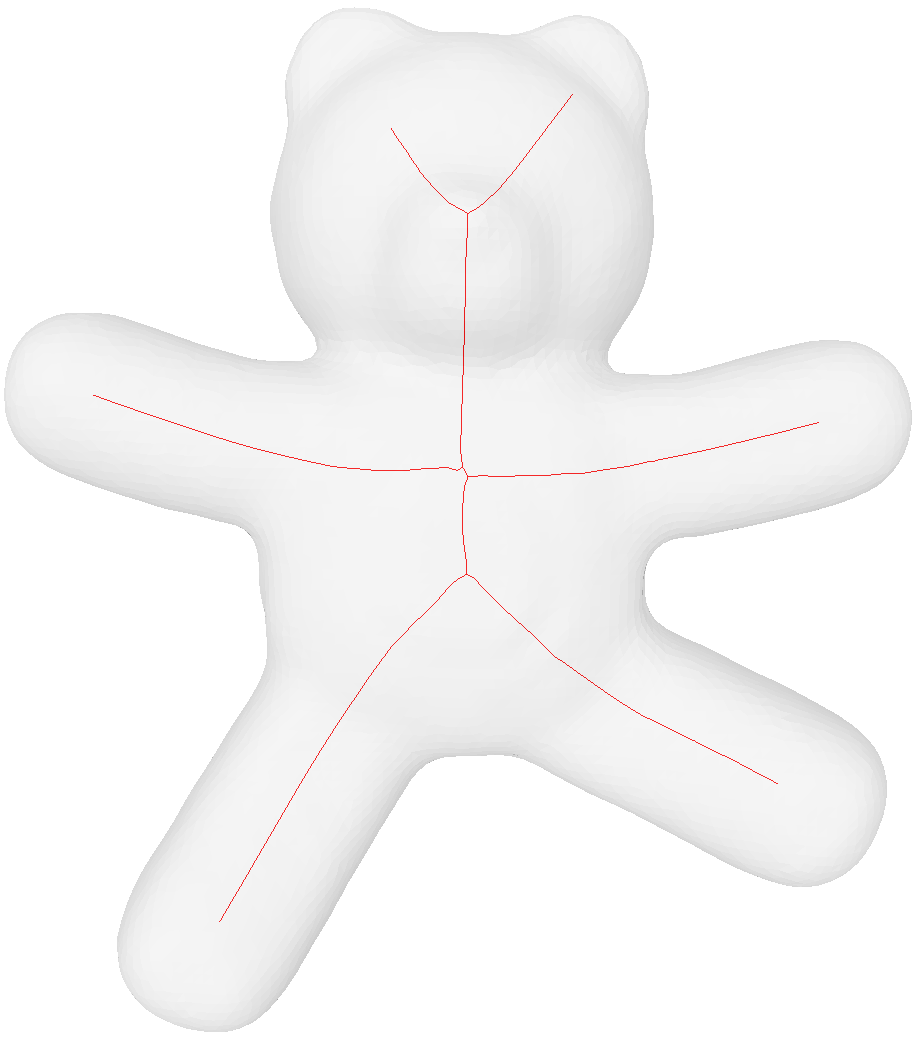}%
\includegraphics[height=0.14\textheight]{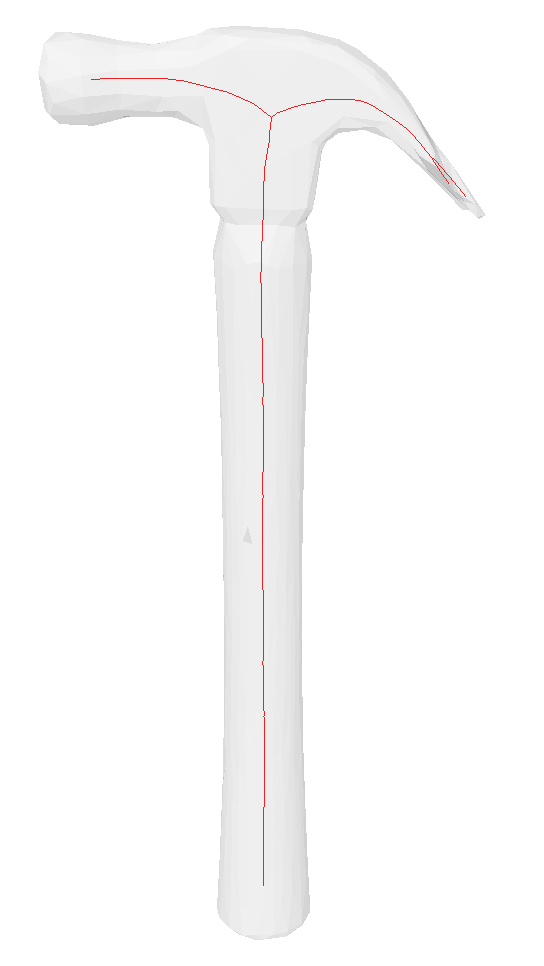}\,%
\includegraphics[height=0.14\textheight]{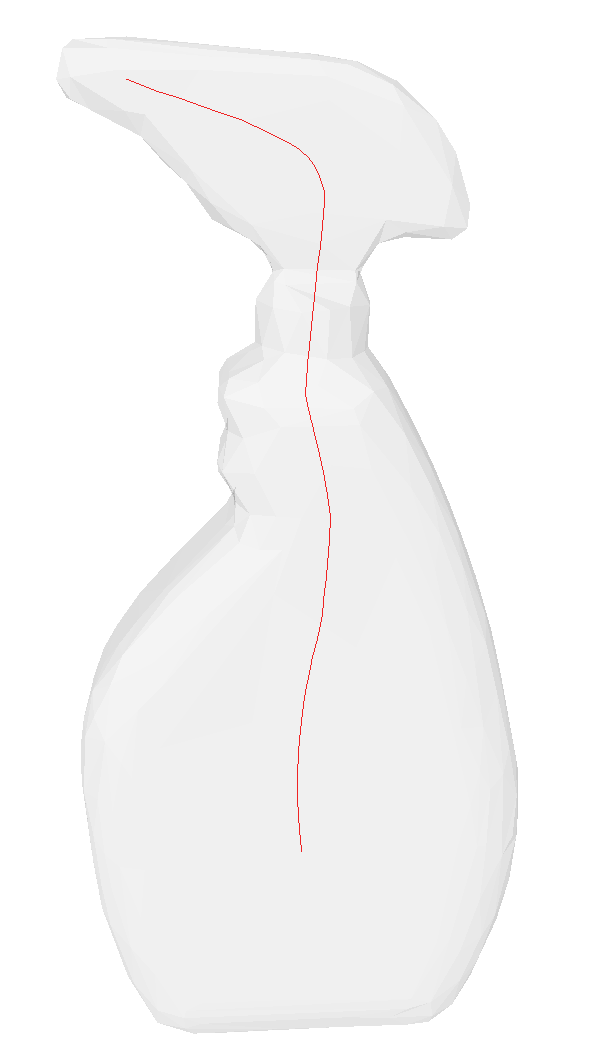}%
\caption{The mean curvature skeleton of several objects.}%
\label{fig:skeleton}%
\end{figure}

Mean curvature skeletons are generated by a contraction-based curve skeleton extraction approach. As input, the 3D mesh of the object is processed in order to retrieve a regularized surface which results in a triangulated object data structure. The set of object surface points are denoted by $O={o_0, \ldots, o_{n-1}}$.
As described in \cite{tagliasacchi2012mean}, the skeleton is build based on iterative mesh contraction via a mean curvature flow approach. Several results are depicted in \autoref{fig:skeleton}.

\begin{figure*}[t]%
\centering
\includegraphics[height=0.17\textheight]{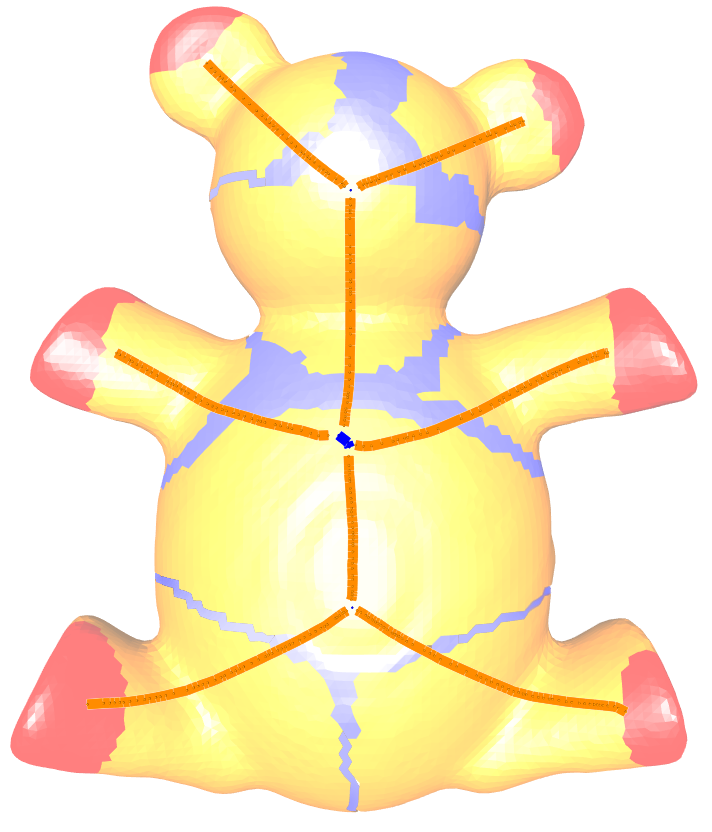}\,\,%
\includegraphics[height=0.17\textheight]{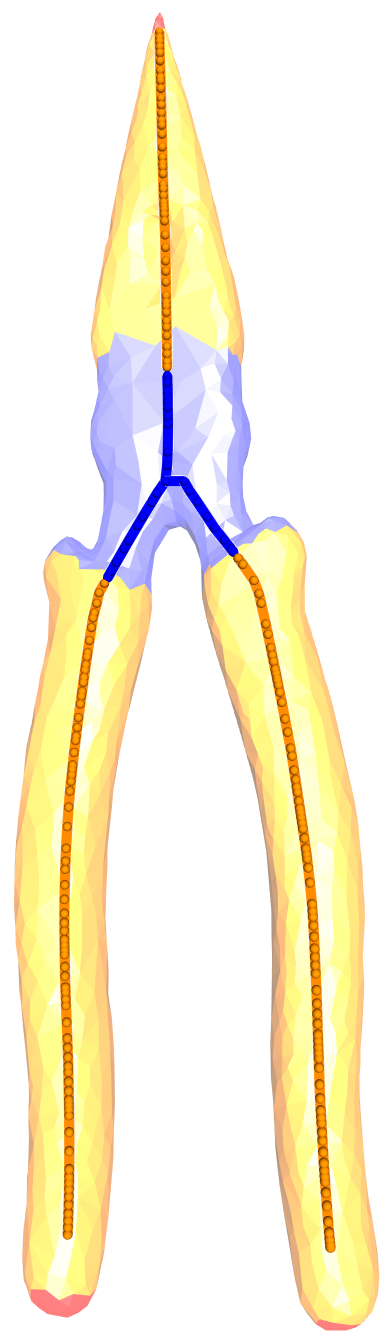}%
\includegraphics[height=0.17\textheight]{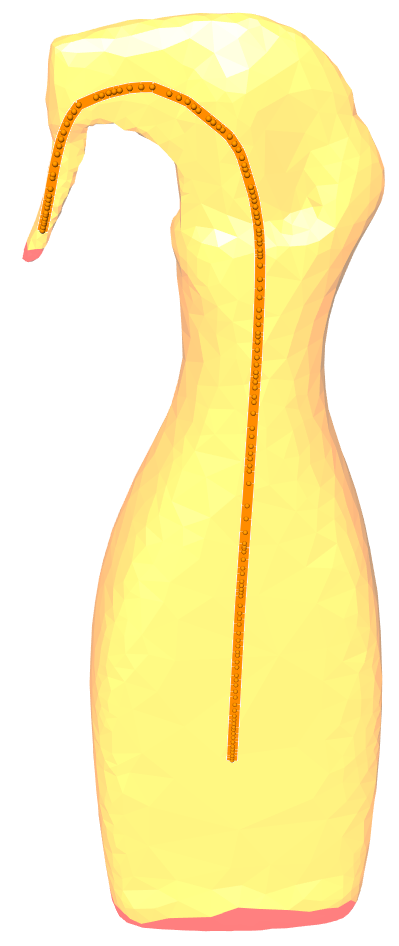}\,\,\,\,%
\includegraphics[height=0.17\textheight]{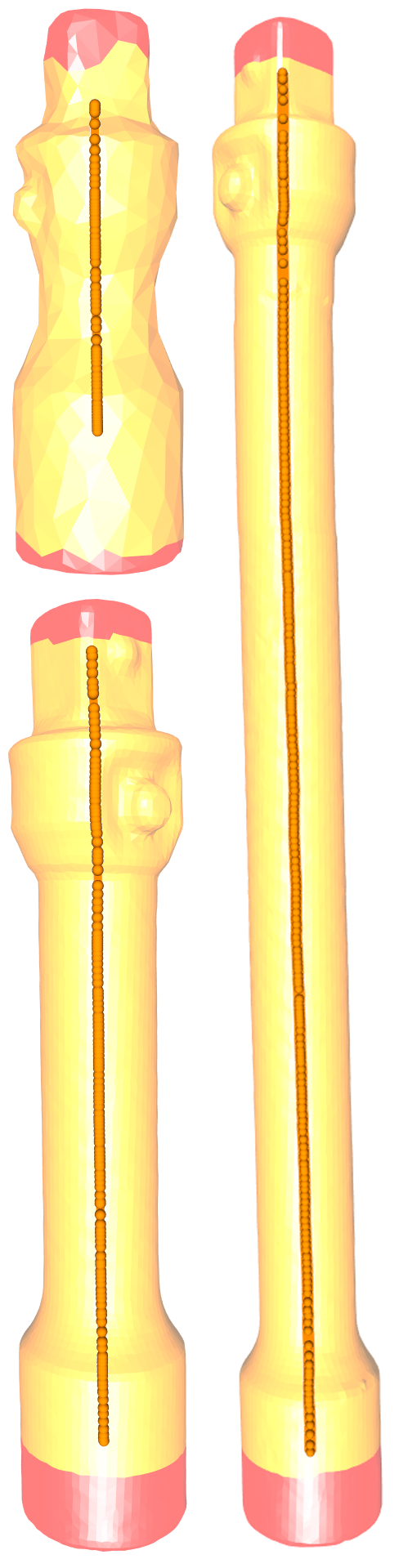}\,\,\,%
\includegraphics[height=0.17\textheight]{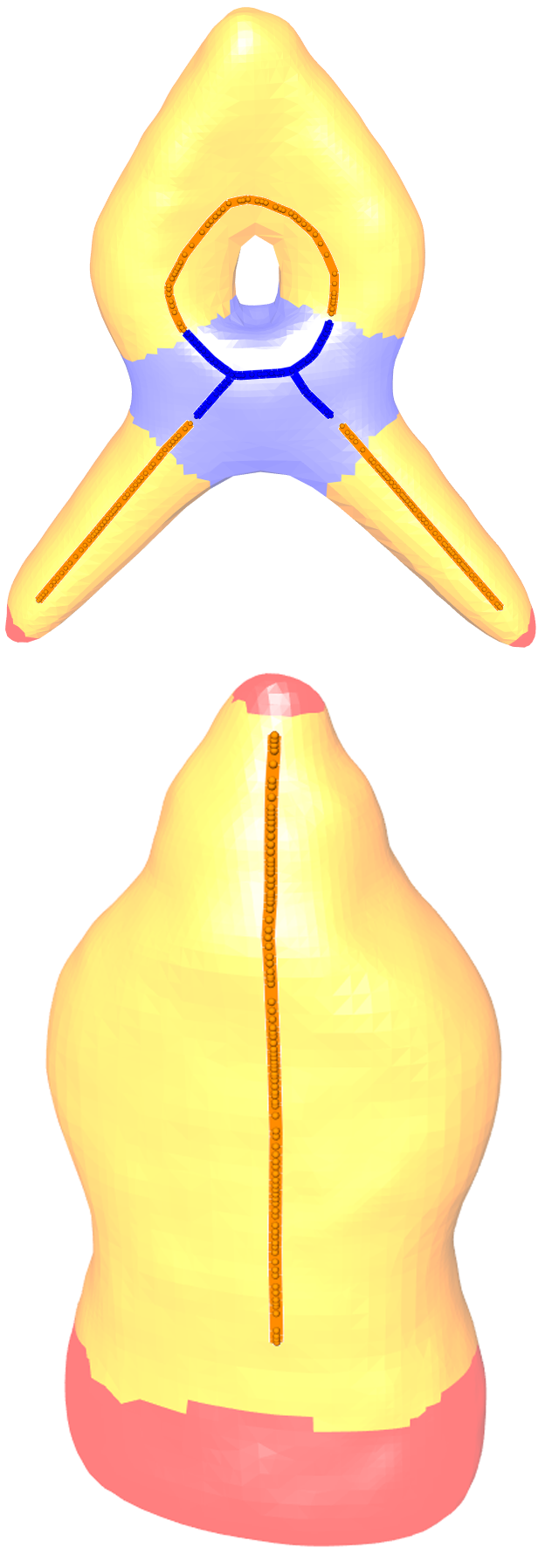}%
\includegraphics[height=0.17\textheight]{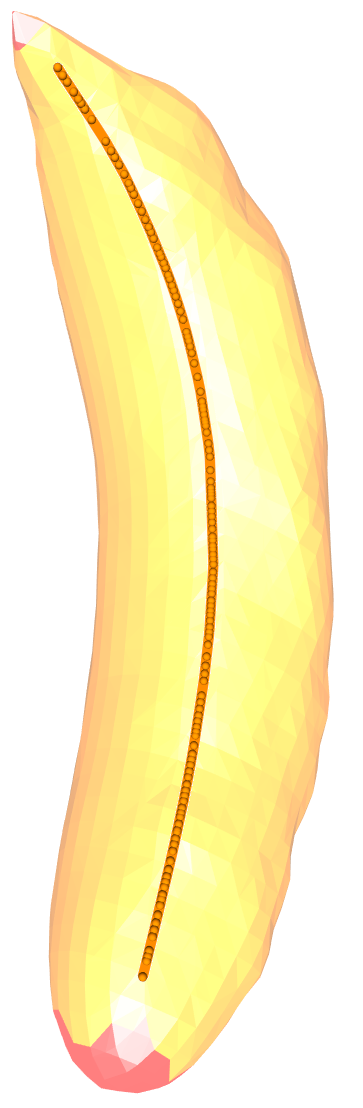}%
\caption{The segmentation of several objects together with the computed skeletons. The surface is colored according to the corresponding skeleton. Branching areas are colored in blue, endpoints result in red color, and the surface associated with connecting segments is visualized in yellow.}%
\label{fig:segmentation}%
\end{figure*}

A resulting skeleton is a graph $S = (V,E)$, in which each vertex $v \in V$ is connected to one or multiple neighbors via edges $e \in E$.
A vertex $v_i = \{s_i, P_i\}$ consist of the 3D position of the skeleton point $s_i \in R^3$ and a set of uniquely associated points on the object surface $P_i \subset O$.
Since all surface points of the object are uniquely associated with a skeleton point, the following relation holds $\sum{|P_i|} = |O|$.
An edge $e = \{v_a, v_b\}$ connects the two vertices $v_a$ and $v_b$.

\subsection{Mesh Segmentation}
\label{sec-segmentation}

The object is segmented based on the skeleton structure in order to generate segments which represent the object's topology. For further processing, each skeleton vertex $v$ is classified according to its connection to neighboring vertices:
\begin{itemize}
	\item \textbf{Branching Vertex:} Such vertices represent a branches or crossings in the skeleton structure. As expressed in \autoref{eq:BranchingVertex}, a vertex $v$ is a branching vertex if there exist more than two edges in the skeleton $S=(V,E)$ containing $v$.
	\begin{equation}
	|\{e \in E: v \in e\}|>2 \Leftrightarrow \text{v is a branching vertex}
	\label{eq:BranchingVertex}
	\end{equation}
	\item \textbf{Endpoint Vertex:} An endpoint vertex $v$ is connected to exactly one other vertex (see \autoref{eq:EndpointVertex}).
	\begin{equation}
	|\{e \in E: v \in e\}| = 1  \Leftrightarrow \text{v is an endpoint vertex}
	\label{eq:EndpointVertex}
	\end{equation}
	\item \textbf{Connecting Vertex:} A connecting vertex $v$ connects to exactly who neighbors as expressed in \autoref{eq:ConnectingVertex}.
	\begin{equation}
	|\{e \in E: v \in e\}| = 2  \Leftrightarrow \text{v is an connecting vertex}
	\label{eq:ConnectingVertex}
	\end{equation}
\end{itemize}

The mesh can now be easily segmented by analyzing the skeleton structure and grouping the skeleton vertices according to their connections. 
A segment $S_i \subset S$ is defined as follows:
\begin{equation}
\begin{split}
	\forall v \in S_i: \text{v is an connecting vertex}\,\,\wedge \\
	\forall e=\{v_a,v_b\} \in S_i: v_a, v_b \in S_i.
\end{split}
\label{eq:Segment}
\end{equation}

The resulting segments $S_i$, contain sub graphs of $S$ consisting of connecting vertices which are enclosed by branching or endpoint vertices.
%
Exemplary segmentation results are depicted in \autoref{fig:segmentation}.

    \section{Part-based Grasp Planning}
\label{sec-planning}

\begin{figure}[tbh]%
\centering
\includegraphics[width=1\columnwidth]{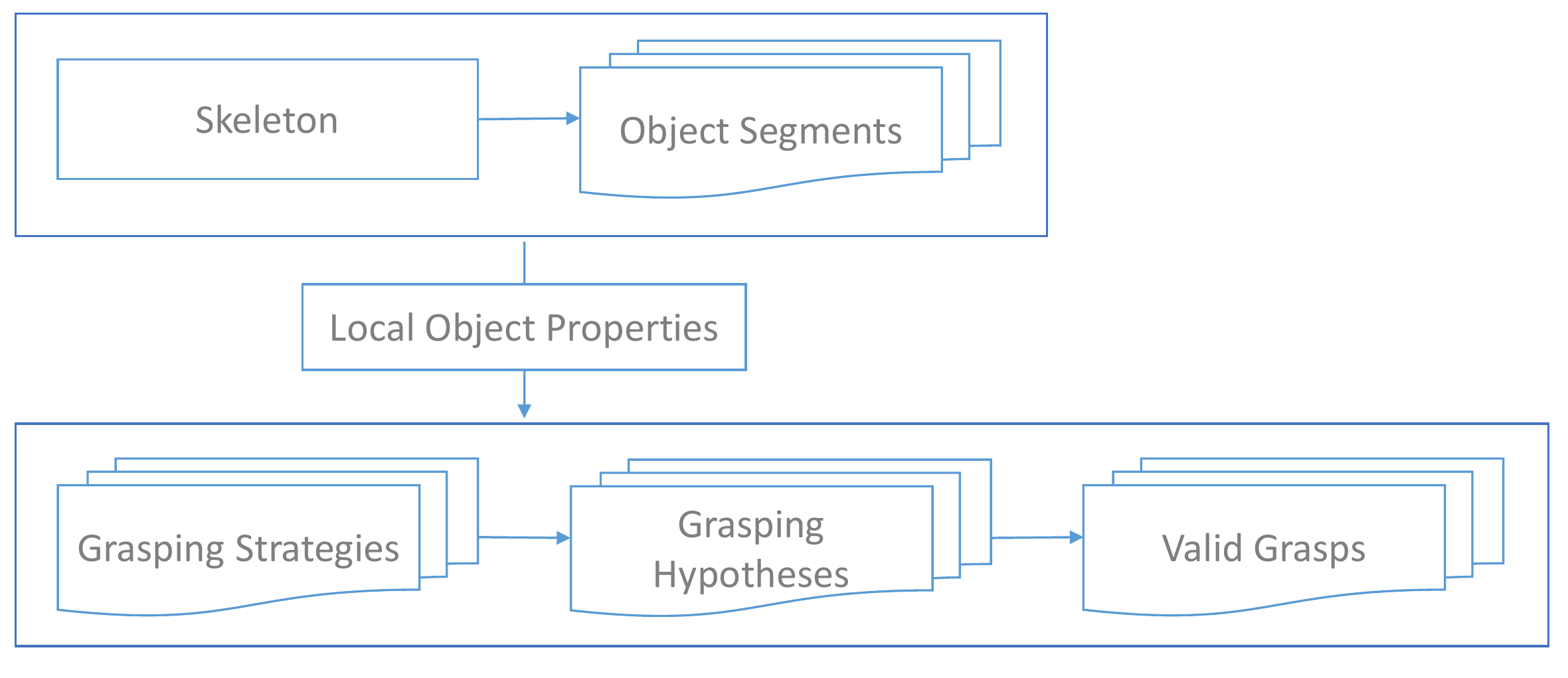}
\caption{The grasp planning process.}%
\label{fig:planning}%
\end{figure}

Grasp planning is performed on the skeletonized and segmented object parts which are analyzed in order to search for feasible grasping poses. 
Therefore, we define several grasping strategies which take into account different local object properties such as the local surface shape or the skeleton structure.
\autoref{fig:planning} depicts an overview of the grasp planning process. 
The grasp planner is shown in \autoref{alg:grasp-planner}. First, a skeleton vertex $v_i$ is selected by processing all skeleton segments and returning all end point vertices, branching vertices, and vertices on connecting segments according to the skeleton vertex distance parameter $d$, which defines the minim distance between two consecutive skeleton vertices. 
Then, we  iterate through all defined grasping strategies and calculate several local object properties. The properties $P$ are used to evaluate if the grasping strategy $gs$ is applicable to the vertex $v_i$. If so, several grasping hypotheses are created. For each hypothesis $h$, a validity check is applied to ensure the correctness of the result. All valid hypotheses are finally stored and returned.

{\SetAlgoNoLine%
\begin{algorithm}[htb]
	\caption{Grasp Planner}
	\label{alg:grasp-planner}
	Input:\\
	\, skeleton $S$, grasping strategies $GS$, vertex dist. $d$\\
	Output:\\
	\, set of valid grasps $G$ \\
\hrulefill \\
	$G = \emptyset$\\
	\While{$(!\text{timeout}() \land \text{verticesAvailable}())$}
	{
		$v_i = \text{nextSkeletonVertex}(S, d)$\\
		\ForAll{$(gs \in GS)$}
		{
			$P = \text{calculateLocalObjectProperties}(v_i, gs, S)$\\
			\If{$(\text{evaluateGraspingStrategy}(gs, P))$}
			{
				$H = \text{generateGraspingHypotheses}(v_i, gs, P)$\\
				\ForAll{$(h \in H)$}
				{
					\If{$(\text{isValid}(h))$}
					{
						$G = G \bigcup \{h\}$
					}
				}
			
			}
		}
	}
	\Return $G$
\end{algorithm}
}

\subsection{Local Object Properties}

To verify that a grasping strategy is applicable, we define several local object properties which can be derived from a skeleton vertex $v = (s,P)$:

\begin{itemize}
\item \textbf{Vertex Type} $P_T$ : Identifies the type of vertex $P_T \in \{connection, endpoint, branch\}$. 
\item \textbf{Grasping Interval} $P_I$: Starting from $v$, $P_I$ includes all outgoing skeleton sub graphs until either a branching or endpoint vertex is reached or a maximum length is travelled on the corresponding graph. Hence, $P_I = \{SG_0, \ldots, SG_s\}$ contains the sub graphs $SG \subset S$ starting from $v$ resulting in $|P_I| = 1$ for endpoint vertices, $|P_I| = 2$ for connecting vertices and $|P_I| > 2$ for branching vertices. Depending on the evaluated strategy, the maximum path length is set to half of the robot hand's width (power grasps) or to the width of one finger (precision grasp).
This information is of interest to determine if a connection segment offers enough space for applying a grasp.
In addition, the curvature $\kappa$ of each point on the grasping interval needs to be limited in order to avoid sharp edges which are difficult to capture in terms of aligning the hand for grasp planning. Hence, a sub graph is cut if the curvature at a specific skeleton point is too high.
The curvature at a skeleton point $s$ is defined as
\begin{equation*}
\kappa(s) = \frac{|| s' \times s'' ||}{||s' ||^3},
\end{equation*}
with the first and second derivatives $s'$ and $s''$ which can be derived by numerical differentiation~\cite{casey1996exploring}. 

\item \textbf{Local Surface Shape} $P_{SS}$: $P_{SS}$ describes the local surface of the object by providing several parameters. First, we reduce the complexity of the analysis of the local object shape by introducing a grasping plane onto which the associated surface points of $P_I$ are projected. The plane is defined by the skeleton point $s$ and the normal parallel to the tangent in $s$.
The considered surface points cover all associated surface points of the skeleton vertices in $P_I$ (see \autoref{fig:grasping-interval}).

\begin{figure}[tbh]%
\centering
\includegraphics[width=0.48\columnwidth]{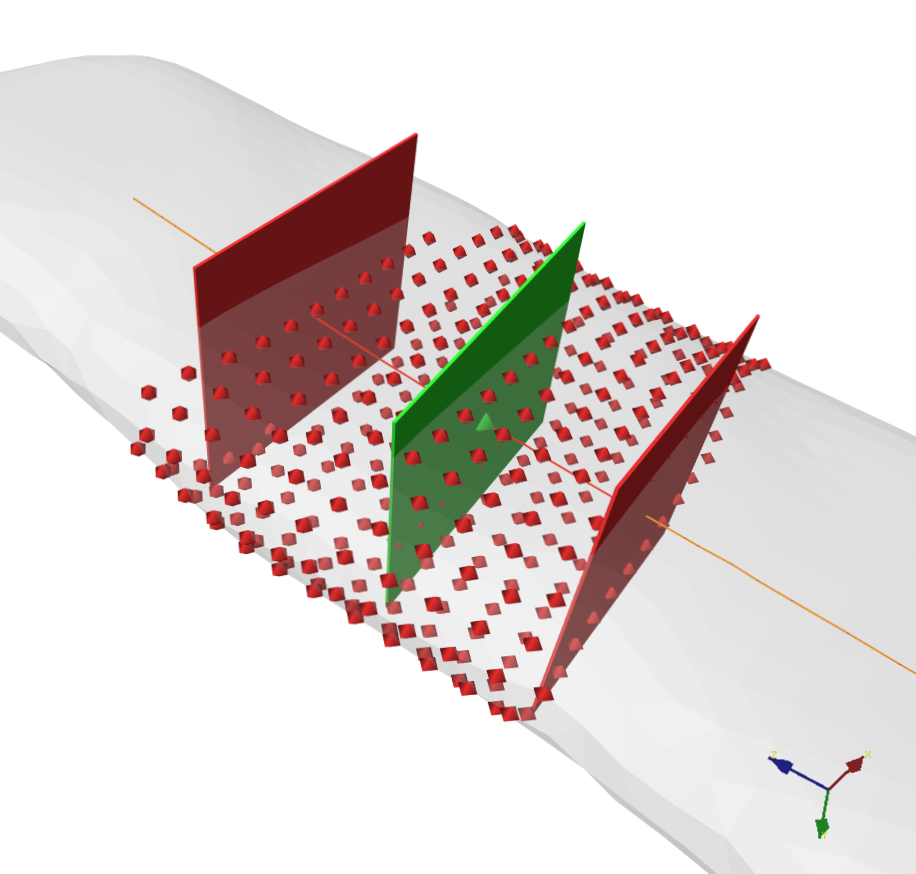}
\includegraphics[width=0.48\columnwidth]{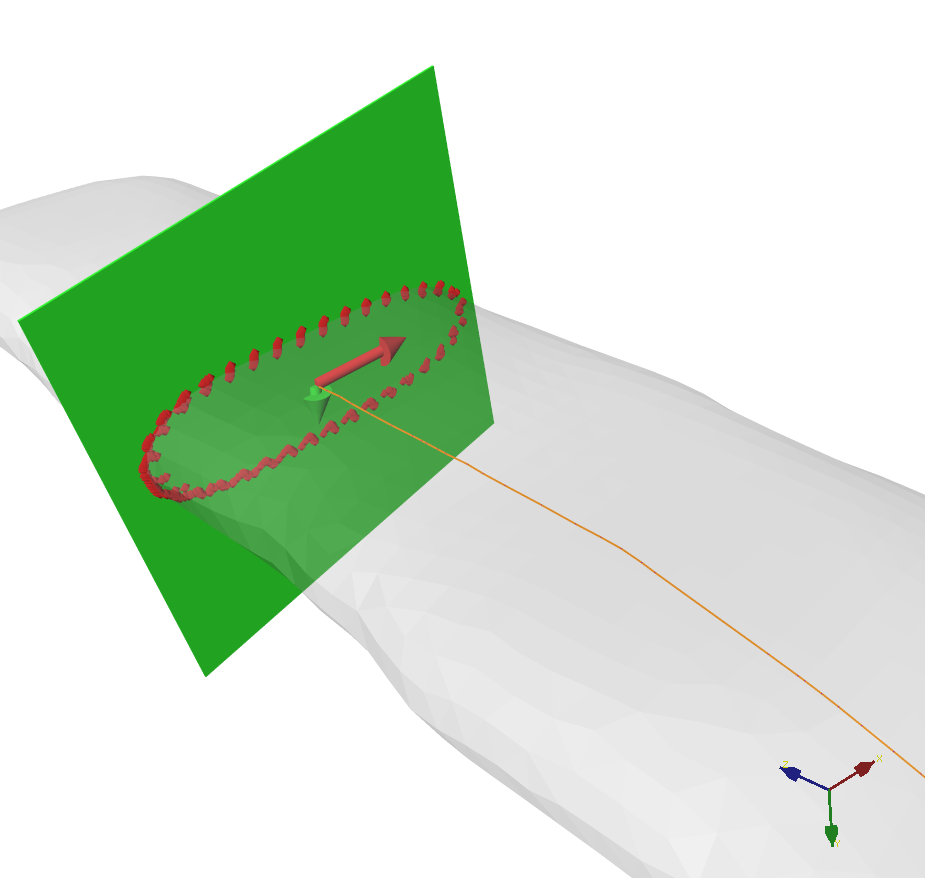}
\caption{Left: Based on a skeleton vertex $v$ (shown in green), a grasping interval $P_I$ is depicted. The red planes define the borders of $P_I$ and the associated surface points of $P_I$ are visualized as red points. On the right, the grasping plane is depicted in green and the projected surface points are shown in red. The corresponding eigenvectors of the projected surface points are visualized as red and green arrows.}%
\label{fig:grasping-interval}%
\end{figure}

The projected surface points are analyzed by applying a principal component analysis to determine the eigenvalues $\lambda_1, \lambda_2$ and the eigenvectors $ev_1, ev_2$. In the following, $\lambda_2$ is used to identify the local thickness of the object.
For further processing, the ratio $r = \frac{\lambda_1}{\lambda_2}$ is calculated and a threshold $t_r$ is used to distinguish between round and rectangular surface shapes. Throughout this work, we use $t_r = 1.2$.
\begin{equation}
shape = \begin{cases}
     round & \text{if } r < t_r \\
     rectangular & \text{otherwise }
   \end{cases}
\label{eq:shape-form}
\end{equation}
Finally, the local surface shape is represented through the 5-tuple $P_{SS} = (\lambda_1, \lambda_2, ev_1, ev_2, shape)$.

\end{itemize}

\subsection{Grasping Strategies}

Our approach allows for setting up a variety of grasping strategies based on the available local and global object information. In the following, we describe several grasping strategies which can be used to generate \textit{precision} and \textit{power} grasps on connection and endpoint parts of the skeleton. To evaluate if a grasping strategy $gs$ can be applied, the local object properties are analyzed as described in \autoref{tab:GraspingStrategies}.
\begin{table*}[t]
  \centering
\begin{tabular}{| R{3.3cm} | M{0.5cm} | M{1.0cm} | M{2.5cm} | M{1.3cm} | M{2.2cm} | M{2.2cm} |}
  \hline			
 \textbf{Grasping Strategy} & \textbf{Nr.} & $\mathbf{P_T}$ 	&  \textbf{Interval Length in} $\mathbf{P_I}$ & \textbf{Shape} &  $\mathbf{\lambda_1}$ &  $\mathbf{\lambda_2}$\\
	\hline
\multirow{2}{3.3cm}{\textit{Precision Grasp on Connecting Segments}} 	& 1a & con. 	& $\geq fingerwidth$					&	round 	
& n/a & $ [pre_2^-,pre_2^+]$\\ \cline{2-7}

																																			& 1b & con. 	& $\geq fingerwidth$					&	rect. 	
& $ [pre_1^-,pre_1^+]$ & $ [pre_2^-,pre_2^+]$\\
\hline	

\multirow{2}{3.3cm}{\textit{Power Grasp on Connecting Segments}}  		&	2a & con. 	& $\geq 0.5 \cdot handwidth$	&	round 
& n/a & $ [pow_2^-,pow_2^+]$\\ \cline{2-7}

																																			& 2b & con. 	& $\geq 0.5 \cdot handwidth$	&	rect. 
& $ >pow_1^-$ & $ [pow_2^-,pow_2^+]$\\
\hline			

\textit{Precision Grasp on Endpoint Vertices } 										& 3 & endpt. 		& n/a 												&	round, rect. 
& n/a & $ [pre_2^-,pre_2^+]$\\
\hline

\textit{Power Grasp on Endpoint Vertices} 												& 4 &  endpt. 		& n/a 												&	round, rect. 	
& n/a & $ [pow_2^-,pow_2^+]$\\
\hline
\end{tabular}
\caption{Grasping strategies are defined according to several local object properties.}\label{tab:GraspingStrategies}
\end{table*}

The grasping strategies can be interpreted as follows:
\begin{enumerate}
\item \textbf{Precision Grasp on Connecting Segments:}
This strategy is applied on a vertex of a connection segment, which means that exactly two skeleton intervals are available in $P_I$. 
The length of each interval has to be at least $fingerwidth$ resulting in an 
accumulated length of the local object skeleton intervals of two times the width of an finger which is reasonable for applying precision grasps. In addition, we distinguish between \textit{round} and \textit{rectangular} shapes of the local object surface.
For \textit{round} shaped objects, we evaluate if the local object thickness, identified by $\lambda_2$ is within the range $[pre_2^-,pre_2^+]$. In case the shape is \textit{rectangular}, we additionally check if the local object length $\lambda_1$ is within $[pre_1^-, pre_1^+]$ in order to bias the decision towards power grasps on objects which provide a reasonable depth. 

\item \textbf{Power Grasp on Connecting Segments:}
Similar to the precision grasp strategy, we analyze the length of both skeleton intervals in $P_I$ for a given vertex of a connection segment.
The length of each interval has to be at least $0.5 \cdot handwidth$ in order to be able to apply a power grasp.
In addition, we distinguish between \textit{round} and \textit{rectangular} shapes of the local object surface.
For \textit{round} shaped objects, we evaluate if the local object thickness, identified by $\lambda_2$, is within the range $[pow_2^-,pow_2^+]$. In case the shape is \textit{rectangular}, we need to exclude small objects and therefore we additionally check if the local object length $\lambda_1$ is larger than $pow_1^-$.
%

\item \textbf{Precision Grasp on Endpoint Vertices:}
This strategy is applied on endpoints of the skeleton structure. Similar to the grasping strategies on connecting segments, the local object shape is interpreted based on the properties of the grasping plane. The length of the local object shape has to be within the range $[pre_2^-,pre_2^+]$ in order to be able to apply a precision grasp.
\item \textbf{Power Grasp on Endpoint Vertices:}
Power grasps are applied on endpoints if the local object length is within $[pow_2^-,pow_2^+]$.
%
%
%
%
\end{enumerate}

 
\subsection{Grasping Hypotheses}
From each grasping strategy several grasping hypotheses are derived and evaluated for correctness (collision-free and force closure) in order to validate the generated grasp.

\begin{figure}[t!]%
\centering
\includegraphics[width=0.4\columnwidth]{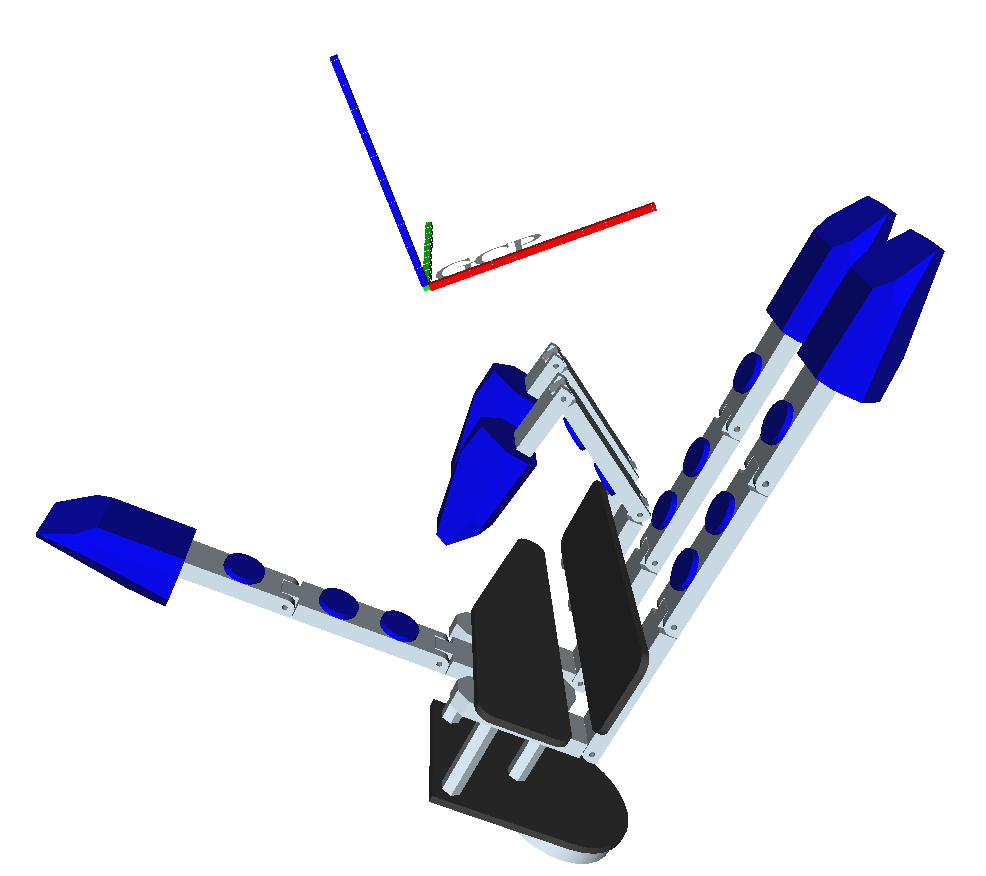}
\includegraphics[width=0.4\columnwidth]{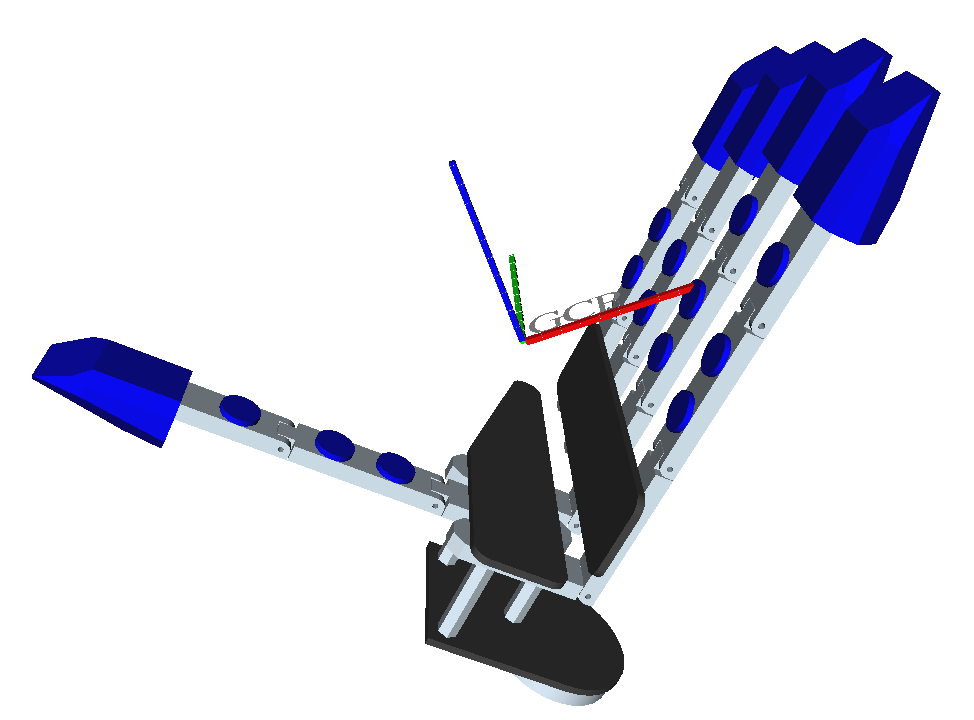}
\caption{The grasp center points of the ARMAR-III hand for precision and power grasps.}%
\label{fig:gcp-armar}%
\end{figure}

\textbf{Grasp Center Points:}
For each hand model, we define a grasp center point (GCP) for precision and power grasps, identifying the grasping center point and the approaching direction \cite{Asfour2008b}. The $GCP_{pre}$ and $GCP_{pow}$ for the robotic hand of ARMAR-III is depicted in \autoref{fig:gcp-armar}. The approach direction is the z-Axis of the depicted coordinate system (visualized in blue).

\textbf{Building Grasping Hypotheses:} For a given skeleton vertex $v$, all valid grasping strategies are evaluated and a set of grasping hypotheses is derived. Therefore, a set of potential approach directions and corresponding hand poses is determined as follows.

\begin{itemize}
\item \textit{Hand Orientation:} The \textit{shape} entry of the Local Surface Shape property $P_{SS}$ results in the generation of different approach directions. 
In case of a \textit{round} local surface, the approach direction is uniformly sampled around the skeleton point $s$. In this case, the approach directions are perpendicular to the skeleton tangent in $s$ for connecting segments and aligned with the skeleton tangent for endpoints. 
If the local object shape evaluates to \textit{rectangular}, four approach directions are built to align the robot hand according to the eigenvectors $ev_1$ and $ev_2$.
In \autoref{fig:approach}, the generated approach directions for one endpoint and one skeleton point on a connection segment are depicted for \textit{round} (left) and \textit{rectangular} (right) local surface properties. In both figures, the approach direction is projected along the negative approach direction onto the object surface. It can be seen that a \textit{round} local surface results in uniformly sampled orientations (in this example, there are eight directions generated for an endpoint, respectively 16 for a skeleton point on a connecting segment). The right figure shows how a \textit{rectangular} local surface results in two approach directions for an endpoint and four approach directions for a skeleton point on a connection segment.
Based on the skeleton points, the set of approach directions and the hand approach vector of the GCP, a set of hand orientations is computed which are used to position the hand in the next step.

\begin{figure}[t!]%
\centering
\includegraphics[width=0.46\columnwidth]{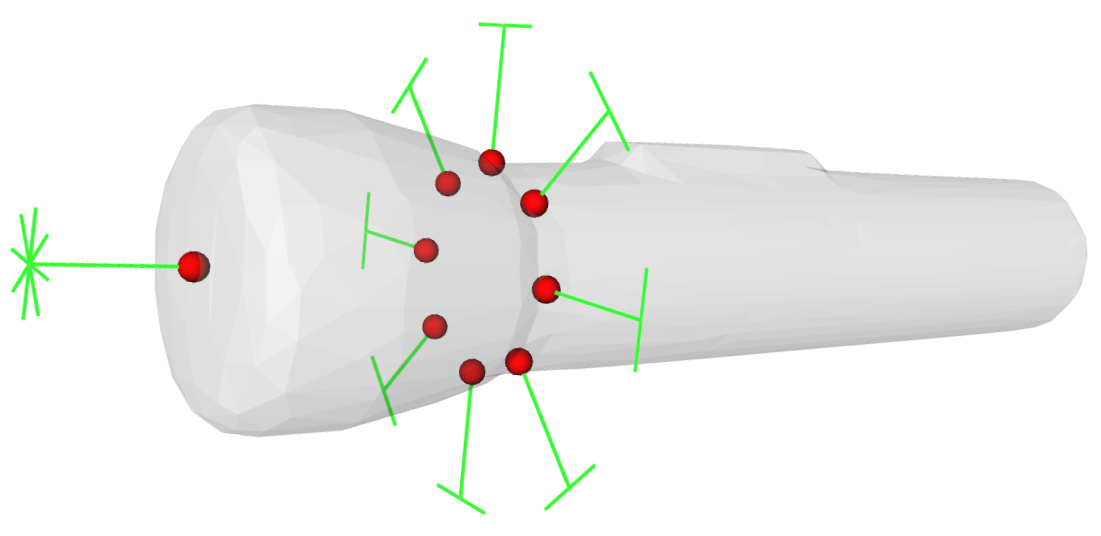}
\includegraphics[width=0.42\columnwidth]{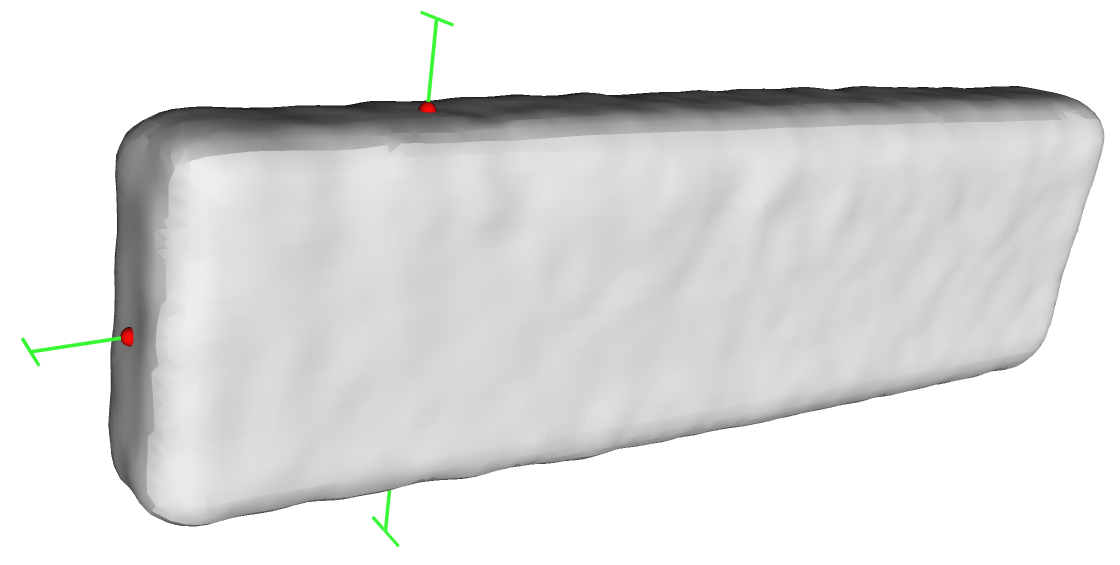}
\caption{The generated approach directions are depicted for an endpoint with local surface properties \textit{round} (left) and \textit{rectangular} (right). In addition, the approach directions for one skeleton point on a connecting segment are depicted for \textit{round} and \textit{rectangular} local surface properties.}%
\label{fig:approach}%
\end{figure}

\item \textit{Hand Position:} The initial position of the hand is derived from the skeleton point $s$. This position is extended to full 6D hand poses by combining it with all computed hand orientations of the preceding step.

\item \textit{Retreat Movement:} To generate a valid grasping hypothesis, the hand model is moved backwards (according to the approach direction) until a collision-free pose is detected. This procedure is aborted if the movement exceeds a certain length. 
\end{itemize}

\begin{figure}[th!]%
\centering
\includegraphics[width=0.28\columnwidth]{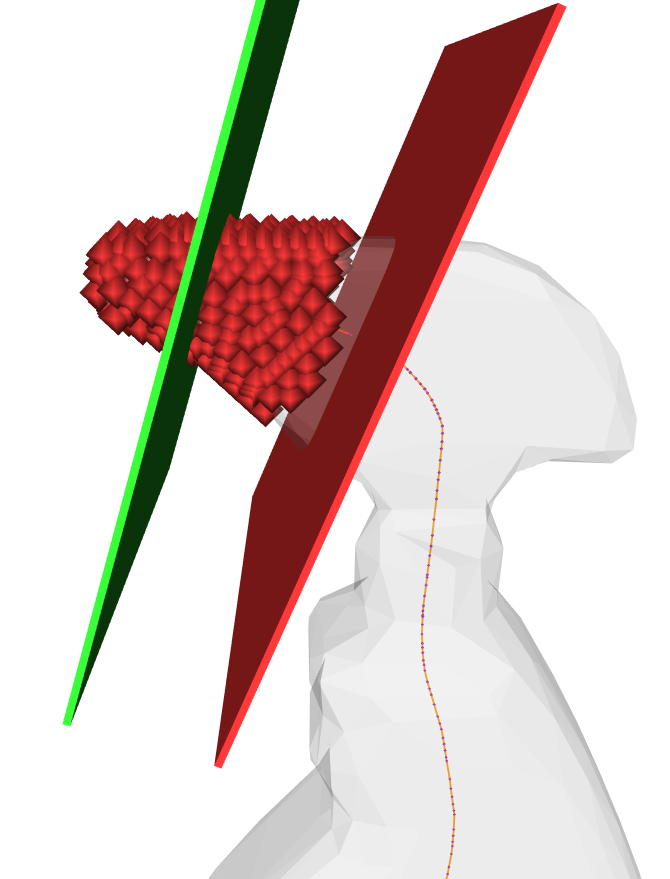}\hspace{2mm}
\includegraphics[width=0.27\columnwidth]{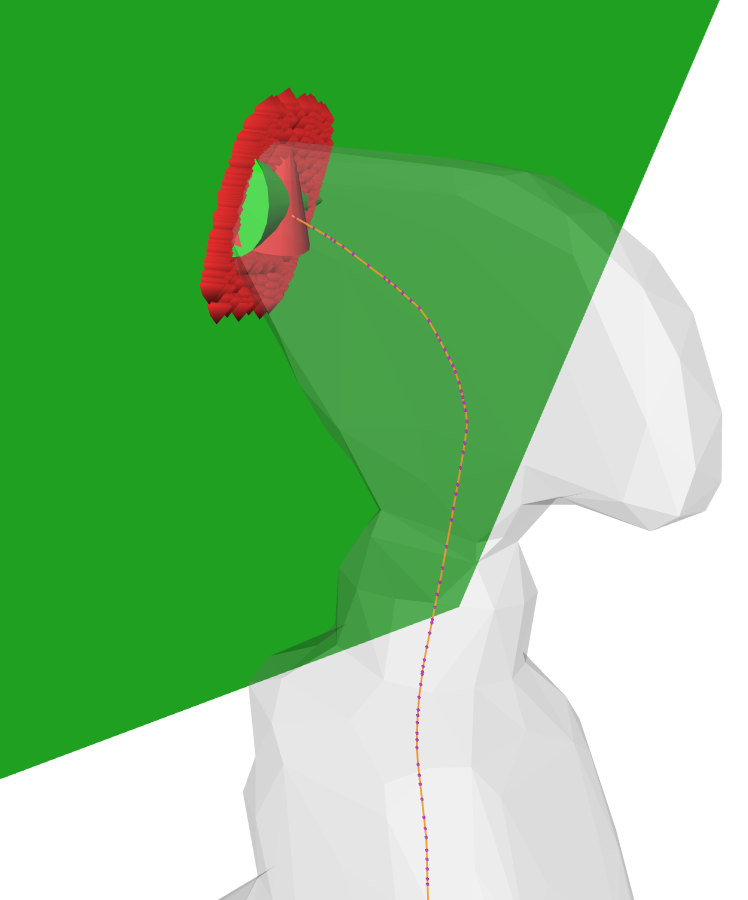}\hspace{1mm}
\includegraphics[width=0.26\columnwidth]{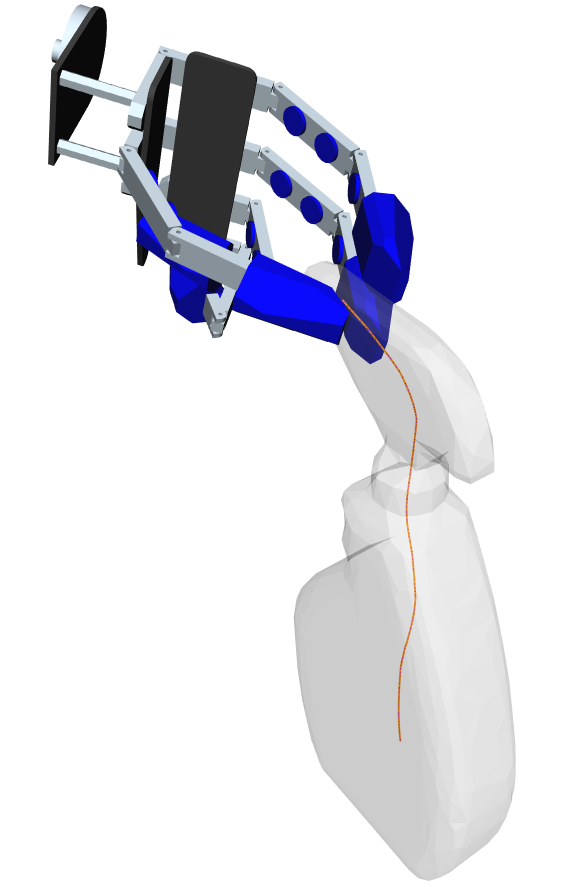}
\caption{(a) The grasping interval together with all associated surface points for a skeleton end point. (b) The grasping plane together with the projected surface points. (c) A generated grasp based on the grasping strategy 1b.}%
\label{fig:spraybottle-endpoint}%
\end{figure}

\begin{figure}[t!]%
\centering
\includegraphics[width=0.19\columnwidth]{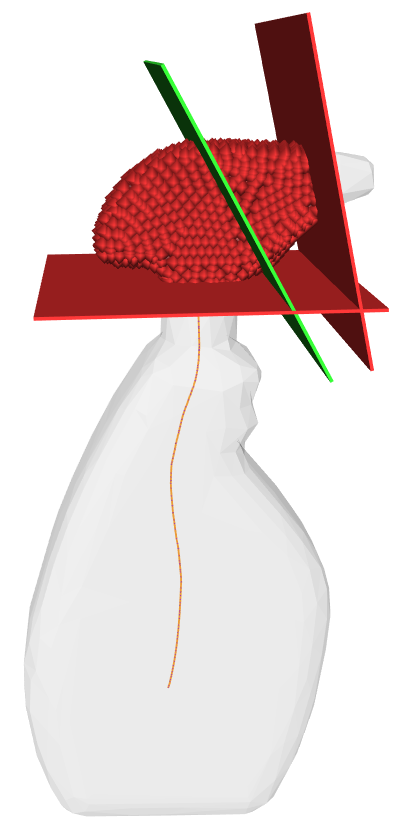}\hspace{5mm}
\includegraphics[width=0.28\columnwidth]{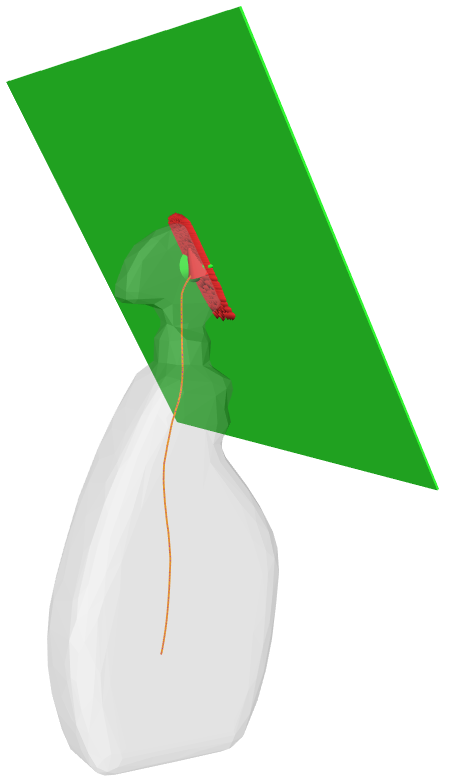}
\includegraphics[width=0.25\columnwidth]{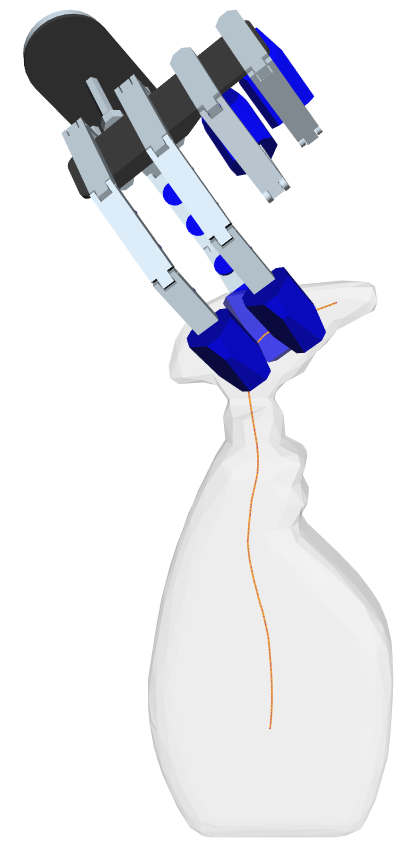}
\caption{(a) The grasping interval together with all associated surface points for a skeleton point on a connection segment. (b) The grasping plane together with the projected surface points. (c) A generated grasp based on the grasping strategy 3.}%
\label{fig:spraybottle-connection}%
\end{figure}

In \autoref{fig:spraybottle-endpoint} and \autoref{fig:spraybottle-connection} the grasping interval, the grasping plane and a generated grasp are depicted for an endpoint vertex and a connection segment respectively.

\begin{figure*}[ht!]%
\centering
\includegraphics[height=0.13\textheight]{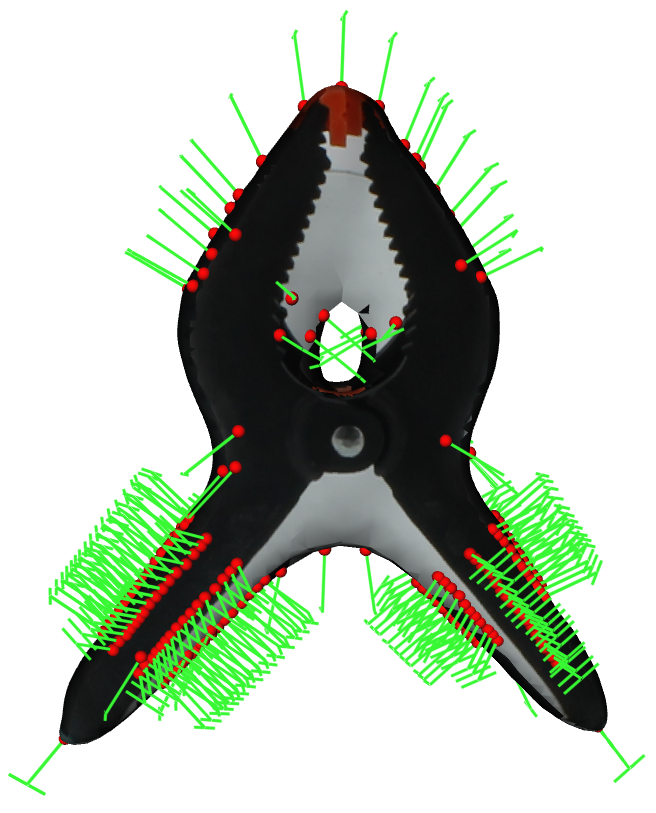}
\includegraphics[height=0.13\textheight]{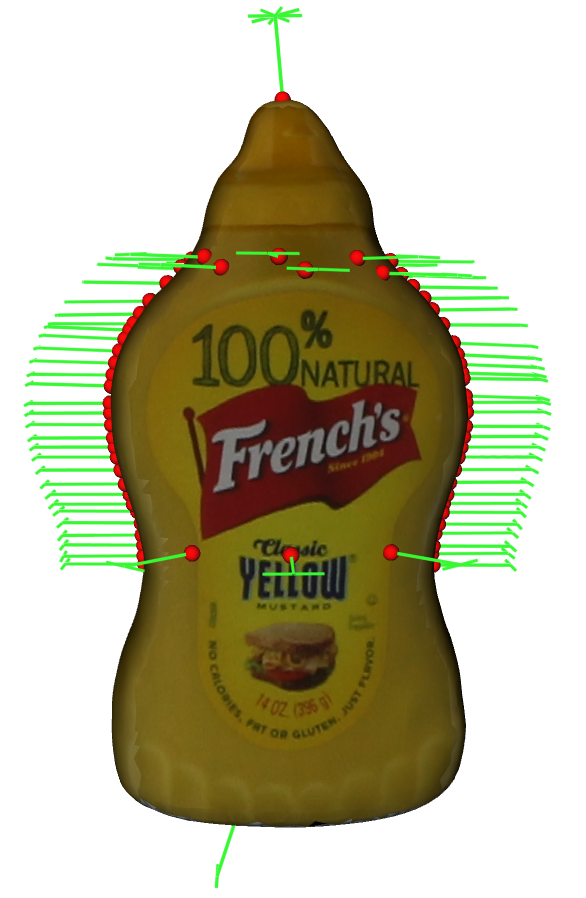}
\includegraphics[height=0.13\textheight]{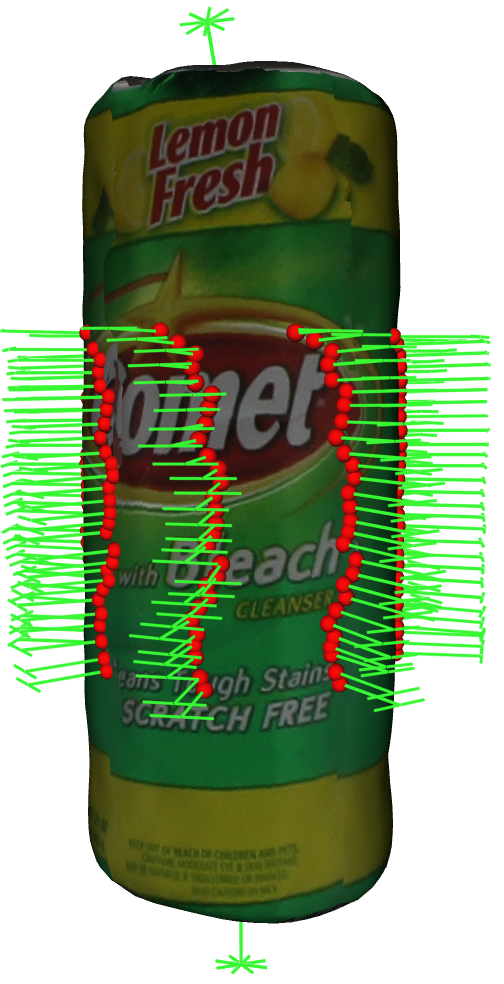}
\includegraphics[height=0.13\textheight]{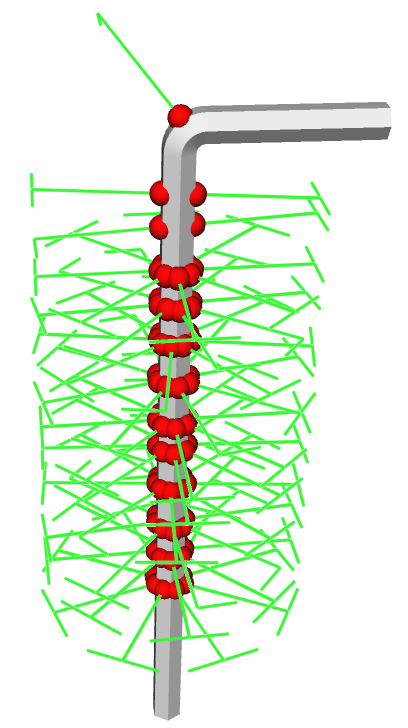}
\includegraphics[height=0.13\textheight]{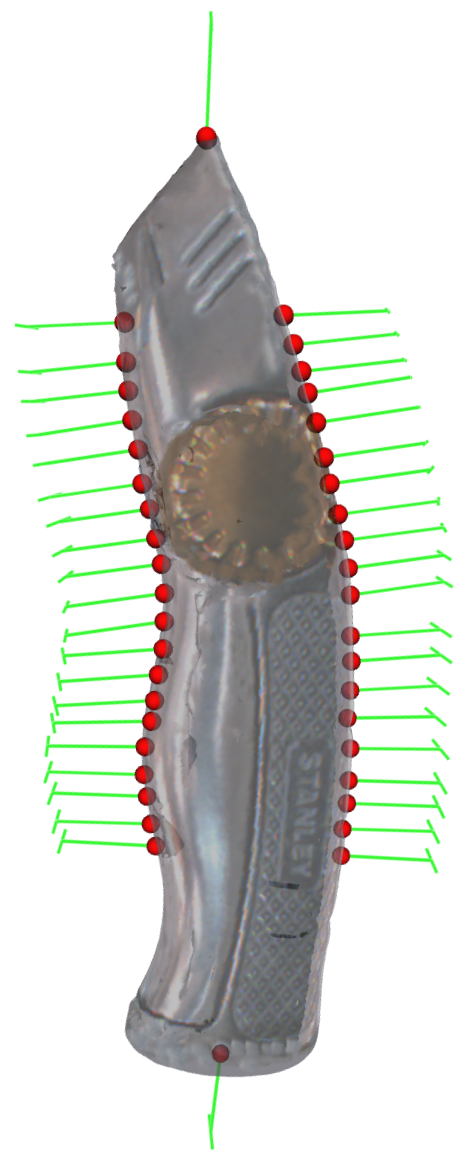}
\includegraphics[height=0.13\textheight]{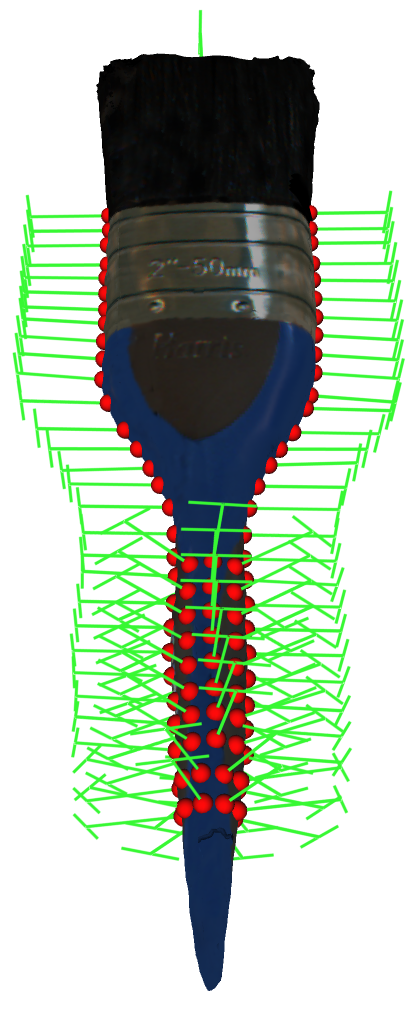}
\includegraphics[height=0.13\textheight]{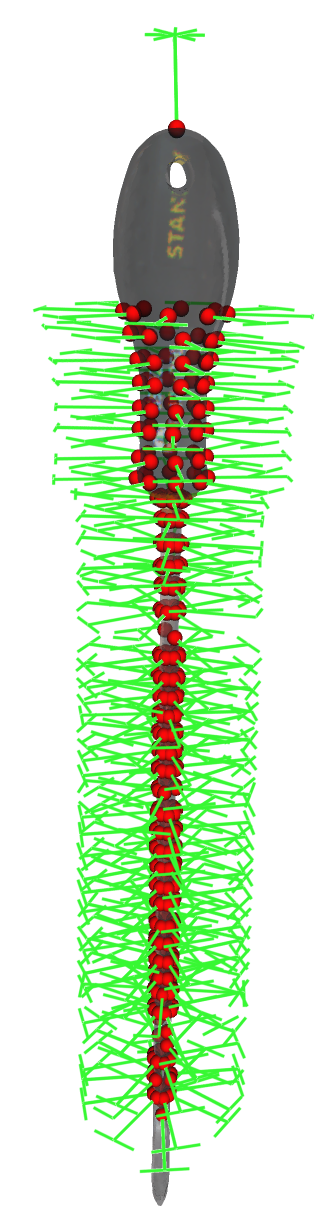}
\includegraphics[height=0.13\textheight]{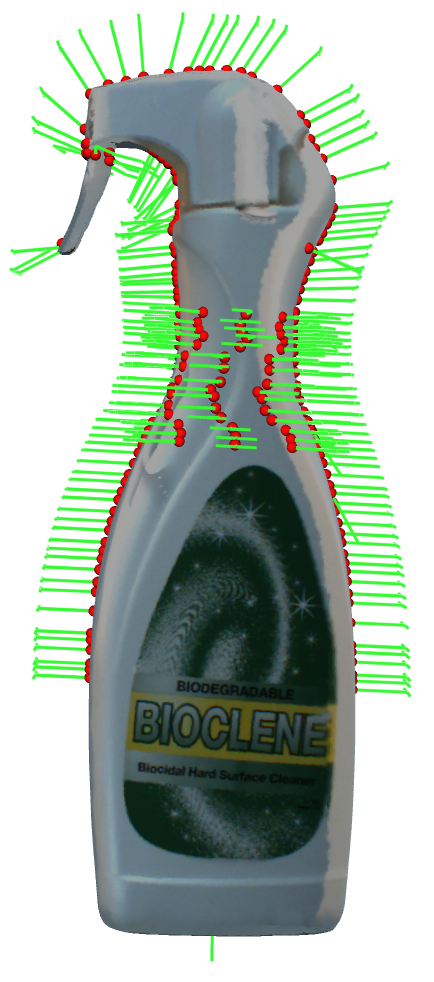}
\caption{Results of the skeleton based grasp planner with several objects of the KIT and YCB object databases. The red dots and the green lines depict the approach movements of the corresponding grasp.}%
\label{fig:results-grasps}%
\end{figure*}

\subsection{Validation of Grasping Hypotheses}

All generated grasping hypotheses are evaluated by closing the fingers, determine the contacts on the object model and by determining if the contacts result in a force-closure grasp. In addition, the quality of the grasp in terms of the grasp wrench space $\epsilon$ value is computed. For grasp stability analysis we employ the methods provided by the Simox library \cite{Vahrenkamp12b}.
All force closure grasps are stored in the set $G$ of valid grasps.
In \autoref{fig:results-grasps} the resulting set of grasps are depicted for several objects of the KIT object model database \cite{Kasper12} and the Yale-CMU-Berkeley (YCB) Object and Model Set \cite{Calli2015}. Note, that the object models were generated based on real-world point cloud data, i.e. no artificial shapes are used for grasp planning. 
The resulting grasps are visualized by projecting the GCP onto the surface according to the corresponding approach direction. The orientation of the grasp is visualized via a green bracket.
A selection of the generated grasps for different hands is additionally shown in \autoref{fig:results-hands}.

    \section{Evaluation}
\label{sec-eval}

The skeleton-based grasp planning approach is evaluated by applying the algorithm on a wide variety of objects of the Yale-CMU-Berkeley (YCB) Object and Model Set \cite{Calli2015} and the SecondHands subset of the KIT object model database \cite{Kasper12}. All object models have been generated by processing point cloud scans of real-world objects. By using such realistic object models, we show that the grasp planning approach can be used under real-world conditions, i.e. without depending on perfectly modeled object meshes.

We compare the results of our approach with a randomized grasp planner \cite{Vahrenkamp12b} which generates power grasps by aligning the approach direction of the end effector to surface normals and evaluates the success by closing the fingers, determining the contacts and evaluating force closure and the grasp quality with the grasp wrench space approach. The planner generates similar results as the approach used in \cite{Diankov2010} and \cite{Kappler2015}.

\begin{figure*}[t!]%
\centering
\includegraphics[width=0.75\textwidth]{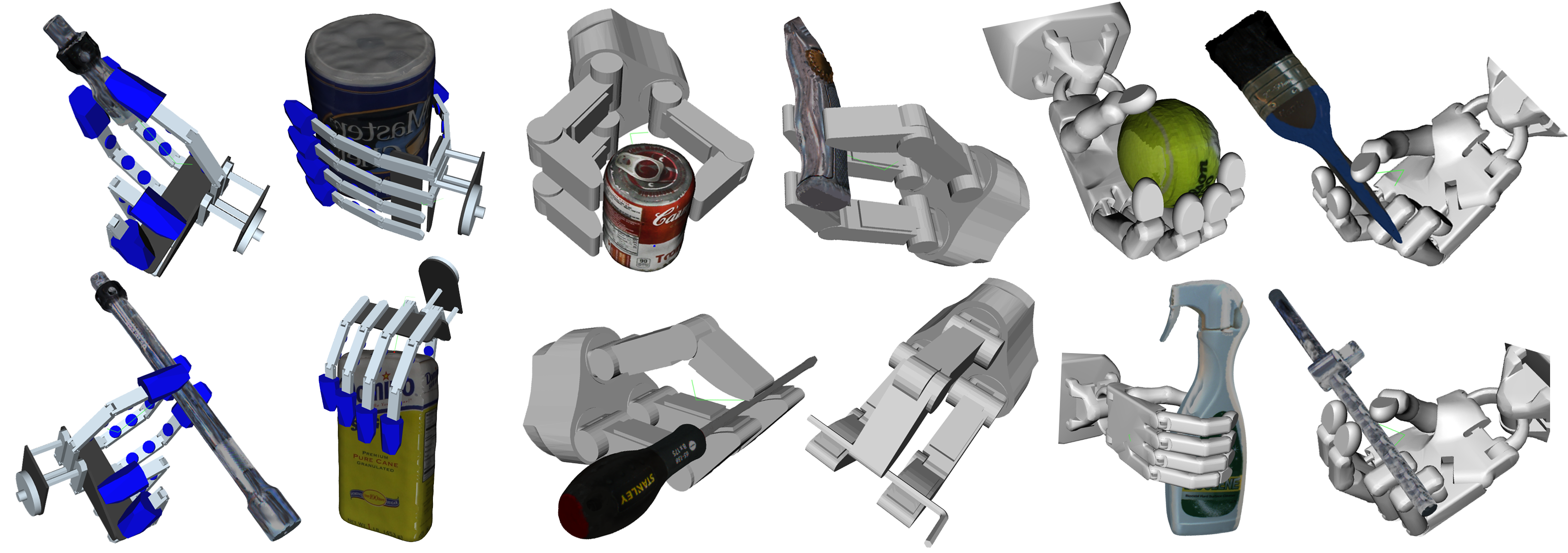}
\caption{Excerpt of the generated grasps for the three hand models that were used in the evaluation.}%
\label{fig:results-hands}%
\end{figure*}

In total, we use 83 objects, 69 from the YCB data set (we exclude all objects with degenerated meshes) and 14 from the SecondHands subset of the KIT object database.
The evaluation is performed with three different robot hand models: The ARMAR-III hand, the Schunk Dexterous Hand, and the Shadow Dexterous Hand (see \autoref{fig:results-hands}). For all hands we define a precision and power preshape with corresponding GCP information. In total, we generated around 4000 valid grasps for each hand.

\subsection{Efficiency and Force-Closure Rate}

The first evaluation concerns the performance of the developed approach in terms of efficiency and force closure rate. In \autoref{tab:eval-time-fc}, the results of the reference planning approach and the skeleton-based grasp planner are shown for the ARMAR-III hand, the Schunk Dexterous Hand, and the Shadow Dexterous Hand. 
The table shows the measured mean values together with the standard deviation over all grasps that were generated on all 83 considered objects. 
The time that is needed for generating a valid grasp directly depends on the complexity of the involved 3D models of the hand since many collision and distance calculations need to be performed when closing the fingers for contact detection.
The comparison shows that the skeleton-based approach outperforms the surface-based planner in terms of efficiency (time needed to generate a grasping pose) and force-closure rate (number of force-closure grasps in relation to all generated grasps). 

When looking at the resulting grasping poses in \autoref{fig:results-hands}, it can be seen that the planned grasping poses are of high quality in terms of what a human would expect how a robot should grasp an object.
Although we cannot provide any numbers on the \textit{human-likeness} of the generated grasps, the underlying algorithm produces good grasps in this sense since the robot hand is aligned with the structure of the object's shape.


\begin{table}[h!]%
\begin{center}
\begin{tabular}{ |c|c|c|c| } 
 \hline
												& Avg. Time & Force-Closure & Robustness \\ 
												& per Grasp & Rate 			& Score $r$\\ 
 \hline
\multicolumn{4}{|l|}{\textbf{Surface-based Grasp Planner}}\\
 \hline
 ARMAR-III Hand 			& $32.29ms$ 		& $57.80\%$  	& $57.17\%$\\   
							& $\pm 23.87ms$ 	& $\pm 28.54\%$ & $\pm 16.18\%$\\
 Schunk Dext. Hand 			& $67.43ms$ 		& $70.78\%$  	& $61.94\%$\\   
							& $\pm 23.22ms$ 	& $\pm 21.72\%$ & $\pm 27.39\%$\\
 Shadow Dext. Hand 			& $90.88ms$ 		& $43.55\%$  	& $55.34\%$\\   
							& $\pm 34.05ms$ 	& $\pm 28.32\%$ & $\pm 18.83\%$\\
 \hline
\multicolumn{4}{|l|}{\textbf{Skeleton-based Grasp Planner}}\\
 \hline
 ARMAR-III Hand 			& $12.19ms$ 		& $95.74\%$  	& $94.23\%$\\   
							& $\pm 3.47ms$ 		& $\pm 11.33\%$ & $\pm 9.52\%$\\
 Schunk Dext. Hand 			& $12.98ms$ 		& $86.69\%$  	& $94.87\%$\\   
							& $\pm 2.72ms$ 		& $\pm 20.50\%$ & $\pm 9.74\%$\\
 Shadow Dext. Hand 			& $28.68ms$ 		& $85.53\%$  	& $76.46\%$\\   
							& $\pm 11.34ms$ 	& $\pm 25.40\%$ & $\pm 20.36\%$\\
\hline
\end{tabular}
\caption{Results of the evaluation.}
\label{tab:eval-time-fc}%
\end{center}
\end{table}
\vspace{-1cm}
\subsection{Robustness}

We evaluate the robustness of the generated grasping information by investigating how inaccuracies in hand positioning would affect the success rate of the grasping process. 
Related to the approach in \cite{weisz2012pose}, we compute a robustness score $r$ for each grasp which indicates how many pose variances within a certain distribution result in a force-closure grasp. 
We create erroneous variances of the grasping pose $p$ by applying a random position and orientation offset to the pose of the grasp. As proposed in \cite{weisz2012pose}, the offset is applied w.r.t. the center of all contacts. The resulting pose $p'$ is then evaluated by moving the hand to $p'$, closing the fingers, detecting the contacts and evaluating if the pose would result in a force-closure grasp. If $p'$ results in an initial collision between hand and object, we count this pose as a failed sample, although there might be numerous grasps which could be executed. To get more detailed information, such situations could be further investigated by applying a physical simulation and considering the approach movement.


The robustness score $r$ is then generated by determining the percentage of force-closure grasps of the total number of displaced samples. In all our experiments we draw 100 samples of displaced grasping poses with a normal distributed error (standard deviation: $10$mm and $5$ degree).

\begin{figure}%
\begin{tikzpicture} 
\begin{axis}[ 
ybar, 
height=5 cm, 
bar width=0.1cm, 
x = 0.38cm,	
enlarge x limits={abs=0.3cm}, 
ymin=0,ymax=0.9,
symbolic x coords={5,10,15,20,25,30,35,40,45,50,55,60,65,70,75,80,85,90,95,100},
xticklabel=$\pgfmathprintnumber{\tick}$\,\%, 
xtick={5, 25, 50, 75, 100},
tick label style={/pgf/number format/fixed}
] 
\addplot[red,fill={red!30!white}] table[col sep=comma,header=false] {./data/plots/niko/armar3-generic-robustness-with-collision.csv}; 
\addplot[blue!80!white,fill={blue!30!white}] table[col sep=comma,header=false] {./data/plots/niko/armar3-robustness-with-collision.csv}; 
\node[] (image) at (rel axis cs:0.2,0.71) {\includegraphics[width=2cm]{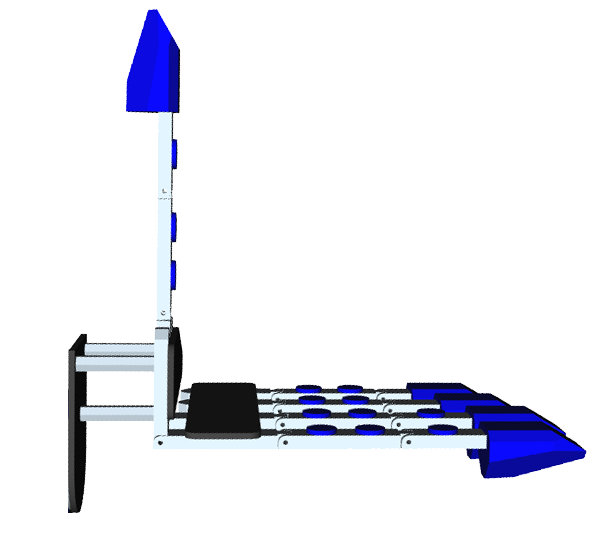}};
\end{axis} 
\end{tikzpicture}
\caption{Robustness histograms for the reference grasp planning approach (red) and the skeleton-based grasp planner (blue) with the ARMAR-III hand. }%
\label{fig:eval-armar3-robustness}%
\end{figure}

\begin{figure}%
\begin{tikzpicture} 
\begin{axis}[ 
ybar,
height=5 cm, 
bar width=0.1cm, 
x = 0.38cm,	
enlarge x limits={abs=0.3cm}, 
ymin=0,ymax=0.8,
symbolic x coords={5,10,15,20,25,30,35,40,45,50,55,60,65,70,75,80,85,90,95,100},
xticklabel=$\pgfmathprintnumber{\tick}$\,\%, 
xtick={5, 25, 50, 75, 100},
tick label style={/pgf/number format/fixed}
] 
\addplot[red,fill={red!30!white}] table[col sep=comma,header=false] {./data/plots/niko/schunk-generic-robustness-with-collision.csv}; 
\addplot[blue!80!white,fill={blue!30!white}] table[col sep=comma,header=false] {./data/plots/niko/schunk-robustness-with-collision.csv}; 
\node[] (image) at (rel axis cs:0.2,0.75) {\includegraphics[width=2cm]{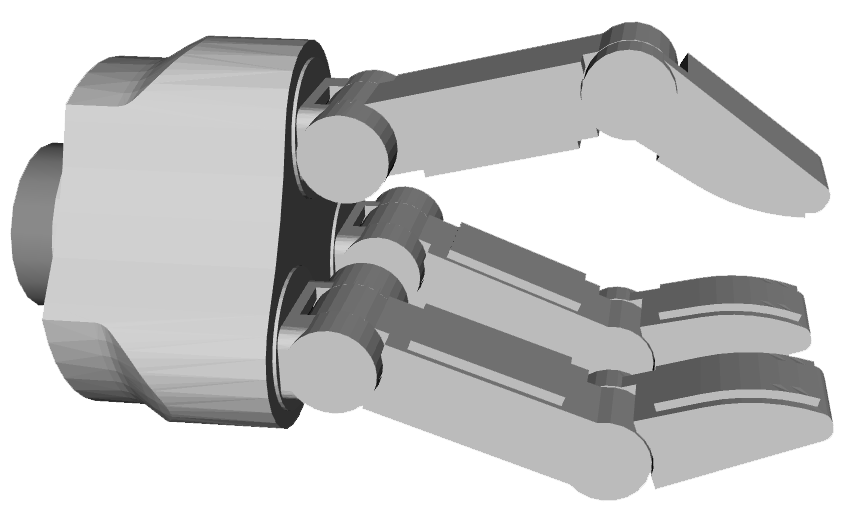}};
\end{axis} 
\end{tikzpicture}
\caption{Robustness histograms for the Schunk Dexterous Hand.}%
\label{fig:eval-schunk-robustness}%
\end{figure}

\begin{figure}[t!]%
\begin{tikzpicture} 
\begin{axis}[ 
ybar,
height=5 cm, 
bar width=0.1cm, 
x = 0.38cm,	
enlarge x limits={abs=0.3cm}, 
ymin=0,ymax=0.4,
symbolic x coords={5,10,15,20,25,30,35,40,45,50,55,60,65,70,75,80,85,90,95,100},
xticklabel=$\pgfmathprintnumber{\tick}$\,\%, 
xtick={5, 25, 50, 75, 100},
tick label style={/pgf/number format/fixed}
] 
\addplot[red,fill={red!30!white}] table[col sep=comma,header=false] {./data/plots/niko/shadow-generic-robustness-with-collision.csv}; 
\addplot[blue!80!white,fill={blue!30!white}] table[col sep=comma,header=false] {./data/plots/niko/shadow-robustness-with-collision.csv}; 
\node[] (image) at (rel axis cs:0.2,0.75) {\includegraphics[width=2cm]{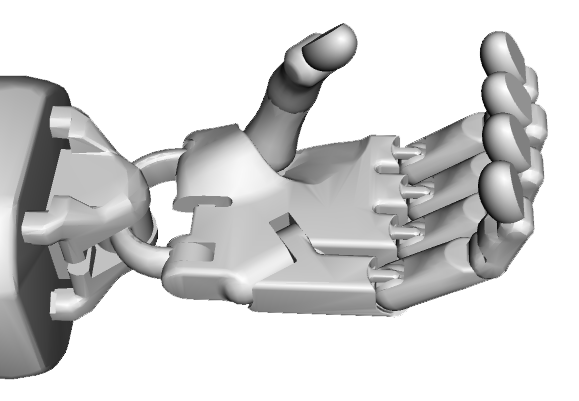}};
\end{axis} 
\end{tikzpicture}
\caption{Robustness histograms for the Shadow Dexterous Hand.}%
\label{fig:eval-shadow-robustness}%
\end{figure}

In \autoref{fig:eval-armar3-robustness}, a histogram of the robustness scores for all generated grasps on all considered objects for the hand of ARMAR-III is shown. The histogram bins on the x axis cover 5\% each and the y axis indicates the absolute percentage of each histogram bin. 
The same histogram is shown in \autoref{fig:eval-schunk-robustness} and in \autoref{fig:eval-shadow-robustness} for the Schunk Dexterous Hand and the Shadow Dexterous Hand respectively.
It can be seen, that the majority of the planned grasps of the skeleton-based approach are robust to disturbances. 

As shown in \autoref{tab:eval-time-fc} in the right column, the robustness score $r$ could be considerably increased for all three hand models when using the skeleton-based grasp planning approach.
It is above $94\%$ for the ARMAR-III and the Schunk Dexterous Hand, which means that $94\%$ of the investigated ill-positioned grasps were leading to force-closure configurations. 
The value for the Shadow Dexterous Hand is lower, which seems to be mainly caused by the fact that the hand kinematics is more complex and the corresponding preshapes provide less room for re-positioning.

The execution of these high-quality grasps with a real robot manipulator would result in higher grasping performance since inaccuracies in perception and/or gripper positioning would still result in a successful grasp.


    \vspace{-0.1cm}

\section{Conclusion}

In this work, we presented a grasp planning approach which takes into account global and local object properties to generate stable and robust grasping information for robotic hands. Global information is gathered by analyzing the object's mean curvature skeleton in order to identify suitable regions for applying a grasp. In addition, local information is used to select the grasp type and to align the hand according to local skeleton and surface properties. 
We showed that the approach is capable of generating high quality grasping information for real-world objects as they occur in object modeling databases such as the KIT or the YCB object DB projects.
We evaluated the approach to a wide variety of objects with different robot hand models and showed that the resulting grasps are more robust in the presence of inaccuracies in grasp execution compared to grasp planners which do not consider local object properties.
In addition, we think that if we compare the results of our grasp planner to the results of other approaches, the generated grasping poses look more natural, i.e. more human like, although we did not perform a qualitative evaluation in this sense. In future work, this aspect will be investigated, e.g. with methods provided in \cite{Borst2005} or \cite{Liarokapis2015}.

    \vspace{-0.2cm}
		
    
    \bibliographystyle{IEEEtran}

     \bibliography{literatur}
    
    %

    
\end{document}